%% file: main.tex
\documentclass[10pt,twocolumn,letterpaper]{article}

\usepackage{cvpr}              

\input{defs.tex}

\usepackage[draft]{changes}
\usepackage{graphicx}
\usepackage{tabularx}
\usepackage{subcaption}
\usepackage{algorithm}
\usepackage{algpseudocode}
\usepackage{multirow}


\definecolor{cvprblue}{rgb}{0.21,0.49,0.74}
\usepackage[pagebackref,breaklinks,colorlinks,citecolor=cvprblue]{hyperref}

\def\paperID{7982} 
\def\confName{CVPR}
\def\confYear{2024}

\title{AAMDM: Accelerated Auto-regressive Motion Diffusion Model}

\author{Tianyu Li\\
Georgia Tech
\and
Calvin Qiao\\
UBC
\and
Guanqiao Ren\\
Beihang University
\and
KangKang Yin\\
SFU
\and
Sehoon Ha\\
Georgia Tech
}

\begin{document}

\makeatletter
\let\@oldmaketitle\@maketitle%
\renewcommand{\@maketitle}{\@oldmaketitle%
    \centering

    \centering
    \vspace{-1em}
        \includegraphics[trim={0 1em 0 0},clip,width=0.95\textwidth]{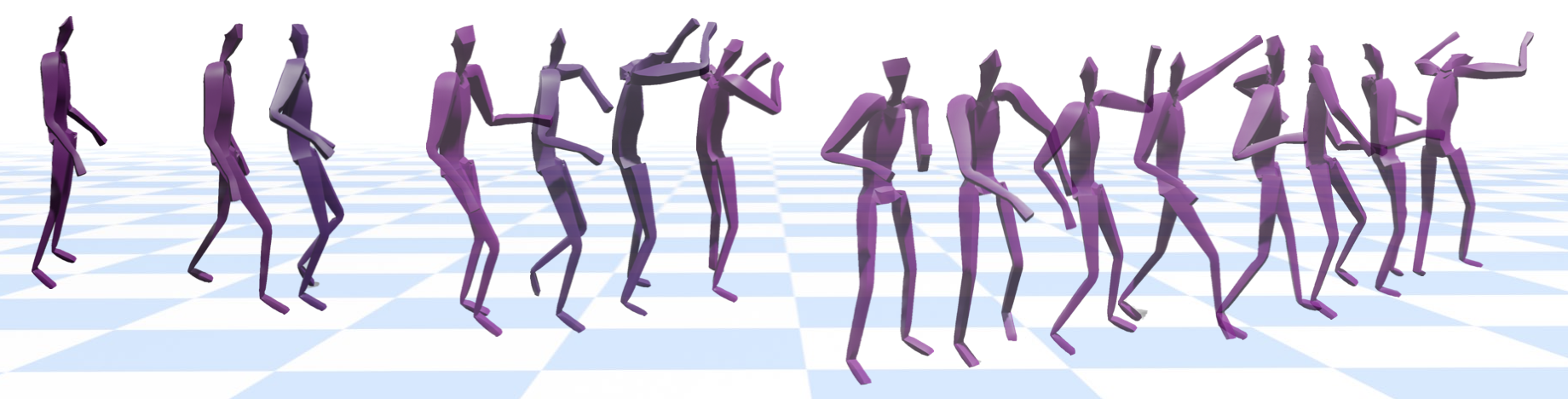}
        \captionof{figure}{\small{We introduce the Accelerated Auto-regressive Motion Diffusion Model (\textit{AAMDM}), a novel framework designed to synthesize diverse and high-quality character motions at interactive rates. }}
    \label{fig:front_page}
    \vspace{0.2cm}
}
\makeatother
\maketitle

\input{paper_text/abstract}
\input{paper_text/introduction}

\input{paper_text/related_work}
\input{paper_text/method}
\input{paper_text/experiment}

\input{paper_text/disscusion_conclusion}

{
    \small
    \bibliographystyle{ieeenat_fullname}
    \bibliography{main}
}

\input{paper_text/appendix}

\end{document}

%% file: defs.tex
\newcommand{\cmt}[1]{}

\newcommand{\bfx}{\mathbf{x}}

\newcommand{\bfz}{\mathbf{z}}

\long\def\ignorethis#1{}

\newcommand{\pctab}{\hspace{0.2in}}



%% file: paper_text/abstract.tex
\begin{abstract}
\vspace{-0.4cm}
Interactive motion synthesis is essential in creating immersive experiences in entertainment applications, such as video games and virtual reality. However, generating animations that are both high-quality and contextually responsive remains a challenge. Traditional techniques in the game industry can produce high-fidelity animations but suffer from high computational costs and poor scalability. Trained neural network models alleviate the memory and speed issues, yet fall short on generating diverse motions. Diffusion models offer diverse motion synthesis with low memory usage, but require expensive reverse diffusion processes. This paper introduces the Accelerated Auto-regressive Motion Diffusion Model (\textit{AAMDM}), a novel motion synthesis framework designed to achieve quality, diversity, and efficiency all together. \textit{AAMDM} integrates Denoising Diffusion GANs as a fast Generation Module, and an Auto-regressive Diffusion Model as a Polishing Module. Furthermore, \textit{AAMDM} operates in a lower-dimensional embedded space rather than the full-dimensional pose space, which reduces the training complexity as well as further improves the performance. We show that \textit{AAMDM} outperforms existing methods in motion quality, diversity, and runtime efficiency, through comprehensive quantitative analyses and visual comparisons. We also demonstrate the effectiveness of each algorithmic component through ablation studies. 
\end{abstract}

%% file: paper_text/introduction.tex
\section{Introduction}

The landscape of interactive motion synthesis, particularly in the realm of video games, has seen a notable expansion. Today's AAA titles boast tens of thousands of unique characters in real-time, all needing to be contextually animated~\citep{holden2018character}. Therefore, the efficiency of motion synthesis has emerged as a critical focus of research in the field of computer animation.
Motion Matching~\cite{LearnedMotionMatching}, a prevalent technique for industry-grade animation, was first developed by UbiSoft for the game \textit{``For Honor''}~\cite{forhonor}. The main objective of Motion Matching (MM) is to identify the most contextually suitable animation in a large dataset based on manually defined motion features. This approach, while capable of yielding responsive high-quality animations, is computationally intensive and scales poorly with respect to the size of the dataset. 

Alternatively, trained neural networks have emerged to reduce the memory footprints and enhance runtime performance. However, these models possess their own challenges, such as unstable convergence at training time and compromised motion quality at testing time. Recently, diffusion-based generative models have revolutionized content creation, thanks to their power to create diverse high-quality content with lean memory demands. However, standard diffusion models are often impractical for time-critical applications, due to their poor run-time performance caused by expensive reverse diffusion processes. 

We introduce the Accelerated Auto-regressive Motion Diffusion Model (\textit{AAMDM}), a novel framework crafted to generate diverse high-fidelity motion sequences without the need for prolonged reverse diffusion. Diffusion-based transition models naturally produce diverse multi-modal motion would be too slow for interactive applications. To overcome this challenge, our \textit{AAMDM} framework mainly adopts two synergistic modules: a Generation Module, for rapid initial motion drafting using Denoising Diffusion GANs; and a Polishing Module, for quality improvements using an Auto-regressive Diffusion Model with just two additional denoising steps. Another distinctive feature of \textit{AAMDM} is its operation in a learned lower-dimensional latent space rather than the traditional full pose space, further accelerating the training process. 

We evaluate our algorithm on the LaFAN1~\citep{harvey2020robust} dataset and demonstrate its capability of synthesizing diverse high-quality motions at interactive rates. Our method outperforms a number of baseline algorithms, such as LMM~\citep{LearnedMotionMatching}, MotionVAE~\citep{ling2020character}, and AMDM~\citep{shi2023controllable}, using various quantitative evaluation metrics. Furthermore, we conduct an analysis on an artificial multimodal dataset. This analysis confirms that our model can successfully capture the multi-modal transition model and is better suited for diverse and intricate motion synthesis tasks. Finally, we perform ablation studies to justify various design choices within our framework. 

In summary, our primary contributions are as follows:
\begin{itemize}
    \item We introduce \textit{AAMDM}, a novel diffusion-based framework capable of generating extended motion sequences at interactive rates. The key idea is to combine the strengths of Denoising Diffusion GANs and Auto-regressive Diffusion Models in a compact embedded space.

    \item We conduct thorough comparative analyses between \textit{AAMDM} and various established benchmarks using multiple metrics for measuring motion quality, diversity, and runtime efficiency. Together with our ablation studies, we provide a deep understanding of our algorithm with respect to alternative prior arts.

    \item We showcase novel high-quality multi-modal motions synthesized from our model, some impossible to achieve by previous methods, such as following a user-controlled root trajectory with diverse arm movements.

\end{itemize}

%% file: paper_text/related_work.tex
\section{Related Work}

\begin{figure*}
\centering
\includegraphics[width=.95\linewidth]{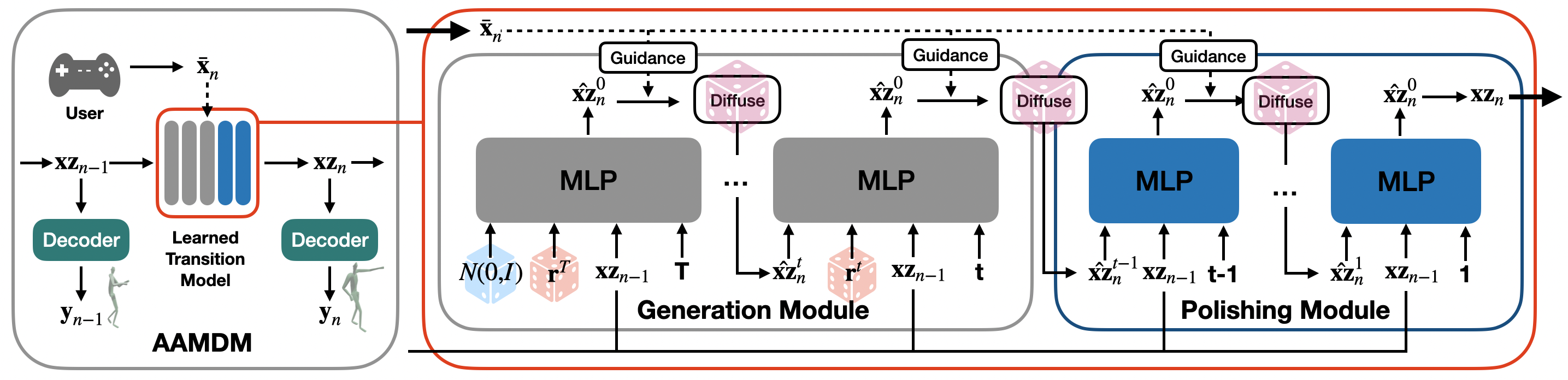}
\vspace{-0.5em}
\caption{\small{Overview of \textit{AAMDM}. \textit{AAMDM} incorporates three pivotal components for better motion quality and faster inference. Firstly, it models transitions within a low-dimensional embedded space $\mathbf{xz}\in\mathbf{XZ}$. Secondly, the framework features a \emph{Generation} module, which employs Denoising Diffusion GANs. This module is responsible for efficiently generating initial drafts of motion sequences. Lastly, a \emph{Polishing} module, which utilizes an Auto-regressive Diffusion Model, refines these initial drafts. 
A full-pose vector $\mathbf{y}_n$ is then reconstructed from the corresponding embedded vector $\mathbf{xz}_{n}$ using the learned decoder $D^{AE}$.
}}
\label{fig:DM3_Overview}
\vspace{-1em}
\end{figure*}

\subsection{Data-Driven Kinematic Motion Synthesis}

The quest to create virtual characters that move naturally stands as a fundamental challenge in computer animation. Graph-based approach structures motion data into a graph and employs search algorithms to retrieve contextually appropriate animations~\citep{arikan2002interactive, kovar2008motion, lee2002interactive, min2012motion, safonova2007construction, hyun2016motion}. It offers high-fidelity motion, but its scalability is curtailed by substantial memory demands and search times.

Statistical methods have been devised to encapsulate motions within numerical models, such as linear, kernel-based, and neural network categories. Linear models represent poses with low-dimensional vectors, but they often fail to encompass the full spectrum of human movement~\citep{howe1999bayesian, chai2005performance, safonova2004synthesizing}. Kernel-based models, including Radial Basis Functions (RBF) and Gaussian Processes (GP), embrace the non-linearity in motion data~\citep{levine2012continuous, wang2007gaussian, grochow2004style, mukai2011motion, mukai2005geostatistical, park2002line, rose1998verbs}. However, these methods are memory-intensive, especially when managing large covariance matrices.

The neural network paradigm has gained prominence for its scalability and efficiency at runtime~\citep{harvey2018recurrent, fragkiadaki2015recurrent, ling2020character, wang2019combining, park2019learning, pavllo2018quaternet, pavllo2020modeling, henter2020moglow, starke2022deepphase,starke2023motion}. Innovative neural architectures have been proposed to better capture motion sequences within datasets, such as those adjusting weights according to a phase variable~\citep{holden2017phase}, employing gating mechanisms~\citep{zhang2018mode}, and extracting periodic latent features~\citep{starke2022deepphase}. Nevertheless, these models predominantly focus on locomotion and character's leg movement, leaving room for broader exploration.

\subsection{Generative Diffusion Model}
Generative diffusion models are a groundbreaking class of algorithms that learn to replicate data distributions through the reverse  of diffusion processes~\citep{song2020denoising,ho2020denoising,DeepUnsupervisedLearning}. In conditional generation scenarios, innovations such as classifier-guided diffusion~\citep{dhariwal2021diffusion} and classifier-free guidance~\citep{ho2022classifier} have been introduced, offering fine-tuned control over the balance between diversity and fidelity. Applications of diffusion models span across image and video synthesis to robotics~\citep{ho2020denoising,song2020improved,ho2022classifier,nichol2021glide,ho2022imagen, hoppe2022diffusion, voleti2022mcvd, janner2022planning, ajay2022conditional}.

Recent adaptations of diffusion models for motion synthesis have been particularly promising, with efforts aimed at generating 3D human motion from textual descriptions~\citep{zhang2022motiondiffuse, tevet2022human, kim2023flame}. Enhancements to these models have come through novel architectural designs~\citep{tevet2022human}, the integration of geometric losses~\citep{tevet2022human}, and the incorporation of physical guidance mechanisms~\citep{yuan2022physdiff}. Additionally, the synthesis of human dance motions from audio signals has been explored, with models using auditory cues to direct the generative process~\citep{alexanderson2022listen, tseng2023edge, ma2022pretrained, dabral2023mofusion}. However, the latency inherent in diffusion models, often taking considerable time to generate brief motion clips, precludes their application in real-time settings. The work of Shi et al.~\citep{shi2023controllable} represents a significant stride towards curtailing inference times through a reduced number of denoising steps.

\subsection{Accelerating Diffusion Model}

The typically slow sampling speeds of diffusion models are primarily attributed to the extensive series of denoising steps required. A range of strategies has been suggested to expedite this process, such as the application of knowledge distillation techniques~\citep{luhman2021knowledge}, the employment of adaptive noise scheduling~\citep{san2021noise}, and the design of single-step denoising distributions as conditional energy-based models~\citep{gao2020learning}. Integrating reinforcement learning with diffusion models has also been proposed to decrease the number of reverse diffusion steps needed~\citep{shi2023controllable}. Nevertheless, such methods have often had to contend with either diminished sample quality or expensive multiple generation steps. The introduction of Denoising Diffusion GANs~\citep{xiao2021tackling} is a notable innovation, integrating the strengths of diffusion models with Generative Adversarial Networks to concurrently address sample quality, generation speed, and mode coverage~\citep{goodfellow2014generative}. In this work, we have employed this technique to enhance the diffusion process for fast motion synthesis.


%% file: paper_text/method.tex
\section{Method}

The architecture of our Accelerated Auto-regressive Motion Diffusion Model (\textit{AAMDM}) is illustrated in Figure \ref{fig:DM3_Overview}. \textit{AAMDM} incorporates three key components: transition in a low-dimensional embedded space, a \emph{Generation} module with Denoising Diffusion GANs for efficient draft generation, and a \emph{Polishing} module with Auto-regressive Diffusion Mode for refining the draft. 

In the following subsections, we will first explain the construction of the low-dimensional embedded space (Section~\ref{sec:embed}). Then, we will describe the foundation of the Polishing (Section~\ref{sec:ADM}) and Generation modules (Section~\ref{sec:ddgan}), namely the auto-regressive diffusion model and denoising diffusion GANs. Next, we will provide the design of the Generation and Polishing modules (Section~\ref{sec:combine}), followed by an explanation of how the sampling procedure is guided to follow user's commands (Section~\ref{sec:guidance}). Finally, we will provide a model representation (Section~\ref{sec:model}).

\subsection{Constructing Embedded Space} \label{sec:embed}
Current learning-based methods for motion synthesis typically target to capture pose transitions in the full-body space, which often complicates learning and violates kinematic constraints intrinsic to the character's morphology. We introduce a compact embedded vector $\mathbf{xz} \in \mathbf{XZ}$ to replace a full-body pose $\mathbf{y} \in\mathbf{Y}$, where $\mathbf{x}$ denotes an engineered feature and $\mathbf{z}$ a learned latent vector.

An autoencoder is employed to learn the optimal embedded space, where an Encoder network $E^{AE}(\mathbf{y})\rightarrow\mathbf{z}$ maps pose vectors to latent vectors, and a Decoder network $D^{AE}(\mathbf{xz})\rightarrow \hat{\mathbf{y}}$ reconstructs poses from the encoded vectors. On top of learned features $\mathbf{z}$, we extract manual feature $\mathbf{x}$ as well. The networks are trained jointly to minimize both the perceptual discrepancy losses, $L^{D,E}_{val}$ and $L^{D,E}_{vel}$, and a regularization loss $L^{D,E}_{reg}$:
\begin{align}
    L^{D,E}_{val} = ||\hat{\mathbf{y}}\ominus\mathbf{y}||+||F(\hat{\mathbf{y}})\ominus F(\mathbf{y})|| \\
    L^{D,E}_{vel} = ||\frac{F(\hat{\mathbf{y}}_0)\ominus F(\hat{\mathbf{y}}_1)
    }{\delta n} - \frac{F(\mathbf{y}_0)\ominus F(\mathbf{y}_1)
    }{\delta n}|| \\
    L^{D,E}_{reg} = ||\bfz||^2_2 \\
    L_{D,E} = w^{D,E}_{val}L^{D,E}_{val}+w^{D,E}_{vel}L^{D,E}_{vel}+w^{D,E}_{reg}L^{D,E}_{reg}
\end{align}
Here, $F$ indicates the forward kinematics function that converts joint rotations into joint positions, and the operator $\ominus$ calculates the difference between two poses. $\mathbf{y}_0$ and $\mathbf{y}_1$ represent two consecutive frames of a motion sequence. $\delta n$ denotes the time interval between frames. $w^{D,E}_{val}, w^{D,E}_{vel}, w^{D,E}_{reg}$ are weights for balancing different loss terms.  
Once we construct the embedded vector space, we can learn an embedded state transition model $S(\mathbf{xz}_{n-1}) \rightarrow \hat{\mathbf{xz}}_{n} $ instead of the full pose transition model $S(\mathbf{y}_{n-1}) \rightarrow \hat{\mathbf{y}}_{n} $. 


\subsection{Auto-regressive Diffusion Model~(ADM) } \label{sec:ADM}

Character animations are intrinsically multi-modal. For a given pose, there may be multiple follow-up poses at the next moment. The transition from $ S( \mathbf { xz } _ { n-1 } ) \rightarrow \hat { \mathbf { xz } } _ { n } $ is essentially a many-to-many mapping. Neural network models that use a Mean Square Error (MSE) based loss to train, such as Learned Motion Matching~\cite{LearnedMotionMatching}, are unable to capture these many-to-many transitions, since MSE losses work on one-to-one mappings. Therefore we employ a diffusion model as our backbone model. Our diffusion model follows the structure of DDPM~\citep{ho2020denoising}. For each forward diffusion step, a small noise vector is added on top of the future embedded vector $ \mathbf { xz } _ { n } $ :
\begin{equation}
    q(\mathbf{xz}^{t}_{n} | \mathbf{xz}^{t-1}_{n}) = N(\sqrt{\alpha^t} \mathbf{xz}^{t-1}_{n}, (1-\alpha^t)I)
\end{equation}

The reverse diffusion phase $p(\mathbf{xz}^{t-1}_{n}|\mathbf{xz}^{t}_{n})$ generates embedded vector $\mathbf{\hat{xz}}_{n}$ by gradually removing the noise on top of $\mathbf{xz}_{n}$. In our setting, the reverse diffusion model $G^{ADM}$ follows the formulation of~\citep{ramesh2022hierarchical, tevet2022human} and directly predicts the embedded vector rather than the added noise as in~\citep{ho2020denoising, shi2023controllable}. The previous vector $\mathbf{xz}_{n-1}$ is used as a condition term:
\begin{equation}
\mathbf{\hat{xz}}^{0}_{n} = G^{ADM}(\mathbf{xz}^t_{n},\mathbf{xz}_{n-1}, t)  
\label{eqn: ADM}
\end{equation}
The predicted $\mathbf{\hat{xz}}^0_{n}$ is then used as a condition $\mathbf{xz}_{n'-1}= \mathbf{\hat{xz}}^0_{n}$ for generating the next $\mathbf{xz}_{n'}$, where $n'=n+1$. 

To ensure high-quality generation over a long horizon, the loss for training $G^{ADM}$ measures the difference between the auto-regressively generated $h$-length embedded vector and the ground truth value.  Specifically, to generate an embedded vector sequence, we start with the trajectory $\bfx\bfz_{ 0:h }$ and add forward diffusion noise. This can be done in an auto-regressive manner using Equation~\ref{eqn: ADM} starting from the initial condition of $ (\bfx\bfz_{0}, \mathbf{xz}^{t_1}_{1}, t_1)$ until $ (\hat{\bfx\bfz}^0_{h-1}, \mathbf{xz}^{t_h}_{h}, t_h)$. The loss is designed as: 
\begin{align}
\label{eqn:ADM_loss}
L^{ADM}_{val} =  ||\hat{\bfx\bfz}^0_{1:h} - \bfx\bfz_{1:h} || \\
L^{ADM}_{vel} =  ||\frac{(\hat{\bfx}^0_{1:h}-\hat{\bfx}^0_{0:h-1})}{h*\delta n} - \frac{(\bfx_{1:h} - \bfx_{0:h-1})}{h*\delta n} || \\
L_{G^{ADM}} = w^{ADM}_{val}L^{ADM}_{val} +w^{ADM}_{vel}L^{ADM}_{vel}
\end{align}
Here $L^{ADM}_{val}$ encourages the reconstruction of the trajectory, and $L^{ADM}_{vel}$ aims to imitate the velocities.

Although this basic diffusion model can produce high-quality samples and achieve improved mode coverage, the sampling process is time consuming primarily due to the iterative nature of diffusion and denoising.

\subsection{Fast Generation via Denoising Diffusion GANs} \label{sec:ddgan}
The Diffusion Model typically involves multiple steps to generate solid predictions. This is based on the assumption that the denoising follows a Gaussian distribution~\citep{DiffGAN}. However, this assumption is only valid when a small amount of noise is eliminated at each denoising step. As a result, it takes numerous steps to generate a high-quality prediction from pure noise. To minimize the number of steps in the reverse process and therefore accelerating the generating process, an alternative approach is to utilize a non-Gaussian multimodal distribution.

Our \textit{AAMDM} utilizes Denoising Diffusion-GANs (DD-GANs) as proposed by \citet{DiffGAN}. This method formulates the reverse diffusion process using a multimodal distribution. It achieves this by parameterizing the reverse diffusion process as conditional GANs. The reverse diffusion generator, denoted as $G^{GAN}$, takes an additional latent variable $\mathbf{r}^t$ as a conditional term, in addition to $(\mathbf{xz}^t_n, \mathbf{xz}_{n-1}, t)$:
\begin{align} 
   \hat{\mathbf{xz}}^0_{n} = G^{GAN}(\mathbf{xz}^t_{n},\mathbf{xz}_{n-1},\mathbf{r}^t ,t).
\end{align}
We use $G^{GAN}(\sim)$ as an abbreviation of $G^{GAN}(\mathbf{xz}^t_{n},\mathbf{xz}_{n-1},\mathbf{r}^t ,t)$, which can be trained by minimizing the KL divergence between two distributions:
$D_{KL}(p(\mathbf{xz}^{t-1}_n|\mathbf{xz}^{t}_n, \mathbf{xz}_{n-1}) || q(\mathbf{xz}^{t-1}_n|\mathbf{xz}^{t}_n, \mathbf{xz}_{n-1})):$
\begin{align}
\label{equ:diffgan_gen}
    L_{G^{GAN}} = -\mathbb{E}_{p(\mathbf{xz}^{t-1}_n|\mathbf{xz}^{t}_n, \mathbf{xz}_{n-1})}[log(D^{GAN}(\sim))].
\end{align}

This objective can then be converted to training a diffusion-step-dependent discriminator network $D^{GAN}(\mathbf{xz}^{t-1}_n, \mathbf{xz}^{t}_n, \mathbf{xz}_{n-1}, t)$  to distinguish if $\mathbf{xz}^{t-1}_n$ is diffused from the original data $\mathbf{xz}_{n}$ or generated fake data $\hat{\mathbf{xz}}^0_{n}$, and the generator is trained to disguise the discriminator. We use $D^{GAN}(\sim)$ as an abbreviation of $D^{GAN}(\mathbf{xz}^{t-1}_n, \mathbf{xz}^{t}_n, \mathbf{xz}_{n-1}, t)$.
\begin{multline}
\label{equ:dis_loss}
    L_{D^{GAN}} \ = \; -\mathbb{E}_{q(\mathbf{xz}^{t-1}_n|\mathbf{xz}^{t}_n, \mathbf{xz}_{n-1})}[log(D^{GAN}(\sim)] \\
     -\mathbb{E}_{p(\mathbf{xz}^{t-1}_n|\mathbf{xz}^{t}_n, \mathbf{xz}_{n-1})}[log(1-D^{GAN}(\sim))]
\end{multline}

DD-GANs offer high sampling speed while maintaining excellent mode coverage and output quality in a single motion step. However, when used in an autoregressive generation setting, we have observed that DD-GANs often lead to unstable training, resulting in deteriorated motion quality. To address this issue, we propose combining ADM and DD-GANs to achieve fast and high-quality sampling.

\subsection{Combining ADM and DD-GANs} \label{sec:combine}
The combination of ADM and DD-GANs is based on the insight that the diffusion process transitions from generating samples from noise at early stages to making small adjustments in the prediction at late stages. To achieve higher quality output, the generation of single motion steps is divided into two sub-steps:  Generation and Polishing. The Generation module utilizes DD-GANs to generate a draft prediction in a few steps, while the Polishing module refines the output from the Generation module using ADM. 

The process begins with a random noise input, $\mathbf{x}\mathbf{z}^T_n \sim N(0,I)$, and the Generation module goes through $T^{GAN}=3$ reverse diffusion steps using $G^{GAN}$ for $\mathbf{x}\mathbf{z}^{T^{AA}-T^{GAN}}_n$. The generated $\mathbf{x}\mathbf{z}^{T^{AA}-T^{GAN}}_n$ is then passed to the Polishing module, where $G^{ADM}$ refines the result using $T^{ADM}=2$ steps. The total number of generation steps, $T^{AA}=T^{GAN}+T^{ADM}=5$. Finally, the generated $\hat{\mathbf{xz}}^0_n$ replaces $\hat{\mathbf{xz}}^0_{n-1}$ for the next step prediction. The Generation Module and Polishing Module are trained separately using Equation~\ref{eqn:ADM_loss}, \ref{equ:dis_loss}, and \ref{equ:diffgan_gen}. For a more detailed sampling and training procedure, refer to the supplementary materials.


\subsection{Motion Control with User Commands} \label{sec:guidance}
To generate the motions that follow the user's commands, for single pose transition, \textit{AAMDM} guides the motion generation process through a guided diffusion method proposed by \citet{rempe2023trace}. Given the user's query $\bar{\bfx}_n$, at each diffusion steps with noise vector $\bfx\bfz^t_{n}$, we perturb the generated vector $\hat{\bfx\bfz}^0_{n}$ to obtain the guided vector $\hat{\bfx\bfz}^{0,*}_{n}$:
\begin{align}
\label{equ:guided_diffusion}
    \hat{\bfx\bfz}^{0,*}_{n} = \hat{\bfx\bfz}^0_{n} - \epsilon \alpha^t \nabla_{\bfx\bfz^t_{n}} J(\hat{\bfx}^0_n, \bar{\bfx}_n).
\end{align}
Here, $J$ is an objective function measuring distance between the generated feature vector and user's query. $\epsilon$ is a step parameter, $\alpha^t$ is the noise parameter in diffusion model.


\subsection{Model Representation} \label{sec:model}
\textbf{Character Representation} The pose vector, denoted as $\mathbf{y}$, captures all the character's pose information in a single frame of the animation. It is defined as $\mathbf{y} = \{\mathbf{y}^t, \mathbf{y}^r, \mathbf{\dot{y}}^t, \mathbf{\dot{y}}^r, \mathbf{\dot{r}}^t, \mathbf{\dot{r}}^r\}$, where $\mathbf{y}^t$ and $\mathbf{y}^r$ represent joint local translations and rotations, $\mathbf{\dot{y}}^t$ and $\mathbf{\dot{y}}^r$ represent joint local translational and rotational velocities, and $\mathbf{\dot{r}}^t$ and $\mathbf{\dot{r}}^r$ represent root translational and rotational velocities. The total dimension of $\mathbf{y}$ is 338. Additionally, we define $\mathbf{x} = \{\mathbf{t}^t, \mathbf{t}^d\}$, where $\mathbf{t}^t \in \mathbf{R}^6$ and  $\mathbf{t}^d\in \mathbf{R}^6$ represents the 2D future trajectory positions and facing direction projected on the ground, 20, 40, and 60 frames in the future local to the character. The latent vector $\mathbf{z}$ has a dimension of 52, and thus $\mathbf{xz} \in \mathbb{R}^{64}$.\\
\\
\textbf{Neural Network Structure}
The encoder network $E^{AE}$, the decoder network $D^{AE}$, ADM generator $G^{GAN}$, DD-GANs generator $G^{GAN}$ and DD-GANs discriminator $D^{GAN}$ are all fully connected neural network. The details are presented in the supplementary materials.

%% file: paper_text/experiment.tex
\section{Experiments}
We conducted a series of experiments to evaluate the performance of the proposed method, \textit{AAMDM}. Firstly, \textit{AAMDM} is quantitatively compared against several baseline methods using different evaluation metrics. Subsequently, we conducted additional experiments on an artificial multi-modal dataset for detailed discussion. Lastly, we performed ablation studies to justify design choices. Overall, the results demonstrate that \textit{AAMDM} can efficiently generate high-quality motions with long horizons auto-regressively. The motions can be seen in the supplementary video.\\
\\
\textbf{Implementation Details}
We implemented our motion generation framework in Pytorch and conducted experiments on a PC equipped with an NVIDIA GeForce RTX 3080 Ti and AMD Ryzen 9 3900X 12-Core Processor.
For all networks, training was performed for $1$M iterations using the RAdam optimizer~\citep{liu2019variance} with a batch size of 64 and a learning rate of 0.0001. We trained the Encoder $E^{AE}$ and Decoder $D^{AE}$ first to construct the embedded vector space $\mathbf{XZ}$, then we trained the Polishing module and the Generation module. Both Polishing and Generation modules were trained with a window size of 10 frames. The total training procedure took around 20 hours. \\
\\
\textbf{Dataset}
We utilized the Ubisoft LaForge Animation Dataset(``LaFAN1'')~\cite{harvey2020robust} for evaluation. LaFAN1 is a collection of high-quality human character animations, encompassing a wide range of motions. Our dataset comprised 25 motion clips from LAFAN1, featuring 100,000 pose transitions, and had a total duration of 26.67 minutes. 

\subsection{Baseline Comparison}

\begin{table*}[h]
\centering

\begin{tabularx}{0.98\textwidth}{ m{2.4cm} m{1.6cm} m{1.5cm} m{1.5cm} m{1.4cm} m{1.6cm} m{1.8cm} m{1.8cm}} 
 \hline 
            & \textbf{DIV} $\rightarrow$ & \textbf{FID} $\downarrow$ & \textbf{FFR} $\downarrow$ & \textbf{FPS} $\uparrow$  & \textbf{TE-UC} $\downarrow$  & \textbf{FID-UC} $\downarrow$ & \textbf{FFR-UC} $\downarrow$\\ [0.7ex] 
 \hline 
 Dataset & 14.533   & 0.000  &  0.113 & N/A & N/A & 0.000 &  0.113\\[0.7ex] 
  \hline 
 \textbf{AAMDM(Ours)} & $\mathbf{11.574}$  & $\mathbf{14.051}$  &$\mathbf{ 0.131}$  &173 & $\mathbf{0.034}$ & $\mathbf{15.367}$ & $\mathbf{0.143}$\\ [0.5ex] 
LMM &5.374  & 49.706 & 0.194 &$\mathbf{812}$ & 0.021 & 55.674 & 0.233\\ [0.5ex] 
 MVAE &7.223  & 22.981 & 0.312  & $\mathbf{703}$ & 0.027& 47.453& 0.349\\ [0.5ex] 
 AMDM5 &8.134  & 18.741 & 0.214 & 192 & 0.054 & 25.943 & 0.256\\ [0.5ex]  
 AMDM200 &$\mathbf{11.165}$  & $\mathbf{12.132}$  & $\mathbf{0.129}$ & 4.72 & $\mathbf{0.012}$& $\mathbf{14.254}$& $\mathbf{0.133}$\\ [0.7ex] 
 \hline
\end{tabularx}
\vspace{-0.5em}
\caption{\small{In our quantitative analysis, we demonstrate that the \textit{AAMDM} framework is capable of generating motions of a quality comparable to that of AMDM200, while significantly outperforming other methods in both random sampling and user control scenarios. Meanwhile, the result also indicates that \textit{AAMDM} is approximately $40$ times faster than AMDM200.
}}
\label{tab: baseline_comparison}
\vspace{-1em}
\end{table*}

We compared \textit{AAMDM} with the following baselines:
\begin{itemize}
    \item \textbf{Learned Motion Matching~(LMM)}: LMM is an interactive motion synthesis method proposed by \citet{LearnedMotionMatching}. Similar to our method, LMM uses an embedded vector space. It comprises three networks: Projector that maps the human input vector $\mathbf{\Bar{x}}$ to the embedded vector $\mathbf{xz}$ for addressing user's command, Decompressor that reproduces the pose vector $\mathbf{y}$ from $\mathbf{xz}$, and Stepper that maps $\mathbf{xz}_{n-1}$ to $\mathbf{xz}_{n}$  for learning the pose transition. Both Stepper and Projector are trained using MSE based loss. Unlike LMM that treats user commands and pose transition separately, \textit{AAMDM} fuses these two requirement using the guided diffusion process. 
    
    \item \textbf{Motion VAE~(MVAE)}: MVAE~\cite{ling2020character} is based on an autoregressive conditional variational autoencoder. Given the current pose, MVAE predicts a distribution of possible next poses, as it is conditioned on a set of stochastic latent variables. The key distinction between \textit{AAMDM} and MVAE is that the former models transitions using a diffusion-based model, whereas the latter employs a VAE.

    \item \textbf{Autoregressive Motion Diffusion Model~(AMDM)}: AMDM~\cite{shi2023controllable} is an autoregressive diffusion model-based framework for motion synthesis. There are three main differences between \textit{AAMDM} and AMDM. First, AMDM accelerates the diffusion process by simply taking fewer reverse diffusion steps, while \textit{AAMDM} leverages DD-GANs. Second, AMDM operates in the full pose space, whereas \textit{AAMDM} learns transitions in an embedded space. Third, AMDM predicts noise at each reverse diffusion step, while \textit{AAMDM} directly predicts the target vector as \citet{ramesh2022hierarchical}. We implemented two versions of AMDM, named AMDM5 and AMDM200, to indicate the use of 5 and 200 diffusion steps, respectively.
    
\end{itemize}

\begin{figure}
    \centering
    \includegraphics[width=0.495\textwidth]{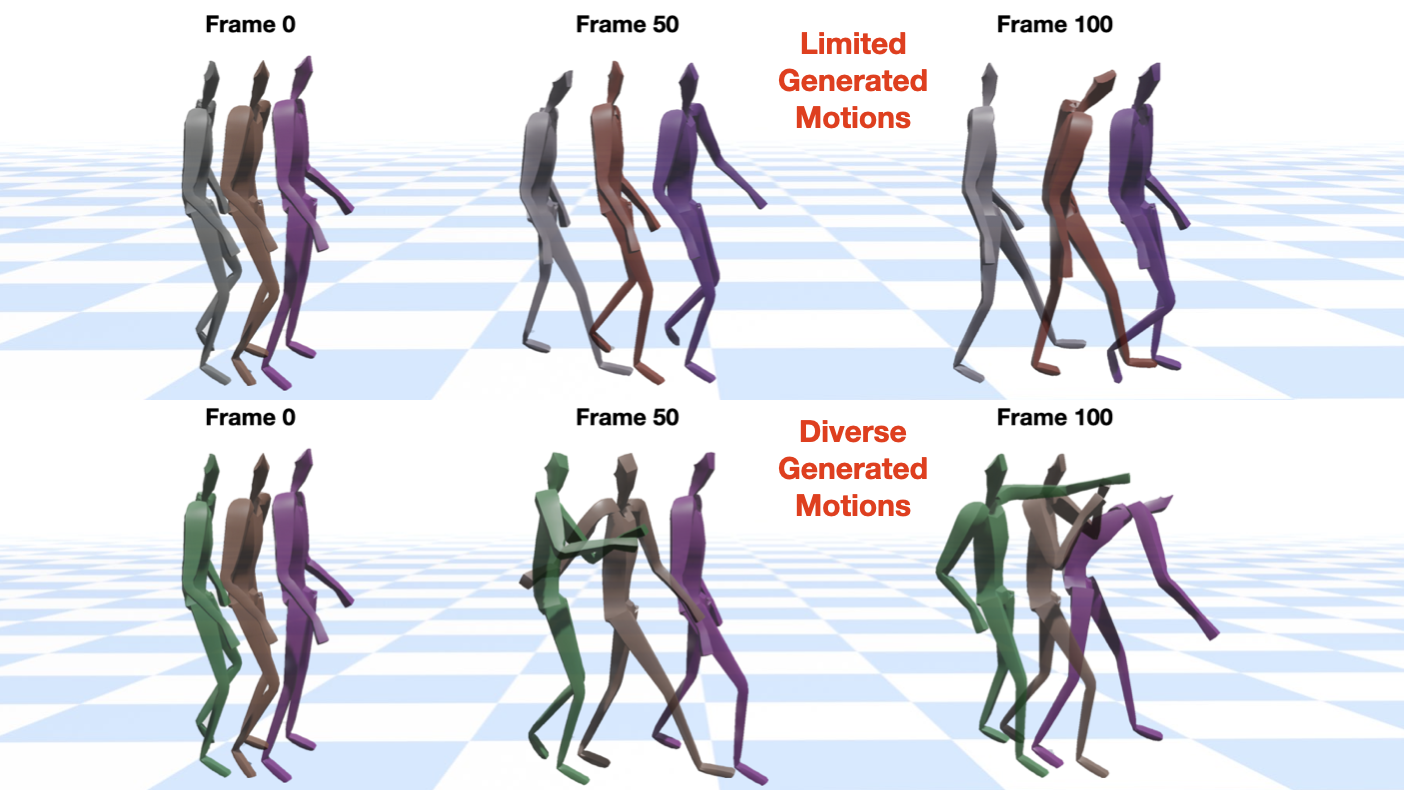}
    \vspace{-1.5em}
    \caption{\small{Comparison between motions generated by LMM (top) and \textit{AAMDM} (Bottom). Starting from a similar character pose, LMM is unable to generate diverse motions while \textit{AAMDM} can reproduce diverse complex motions.}}
    \label{fig:diverse_example}
    \vspace{-2em}
\end{figure}

\subsubsection{Evaluation of Random Motion Synthesis}
We first evaluated the performance of these methods in random motion generation over the following metrics:
\begin{itemize}
\item \textbf{Diversity~(DIV)}: Diversity measures the distributional spread of the generated motions in the character pose space. This metric, adopted from several previous works~\citep{tevet2022human, tseng2023edge,crossloco,ACE}, assesses how well the generated motion matches the distribution of the ground truth dataset. We follow the implementation used in MDM \citep{tevet2022human}, computing Diversity using 1,000 frames from each generated motion clip. A good Diversity score should be close to that of the motion dataset.


\item \textbf{Frechet Inception Distance~(FID)}: FID evaluates the difference between the distributions of generated and ground truth motions. FID serves as an indicator of the overall quality of generated motions in many prior works~\citep{tevet2022human, tseng2023edge}.


\item \textbf{Footskating Frame Ratio~(FFR)}: FFR quantifies the realism of generated motion, particularly focusing on foot-ground contact. We measured foot skating artifacts as described in \citet{zhang2018mode}. A lower FFR score indicates better physical plausibility of the generated motions.

\item \textbf{Frames Per Second~(FPS)}: FPS is a measure of the efficiency of motion generation methods in creating new frames. Higher FPS values indicate faster frame generation rates, essential for interactive applications.
\end{itemize}

The qualitative results are summarized in Table \ref{tab: baseline_comparison}. Notably, \textit{AAMDM} achieve similar performance as AMDM200 while surpasses the other baselines in all motion quality metrics with more than 40 times faster than AMDM200. This demonstrates \textit{AAMDM}'s capability to efficiently generate high-quality character animations.

LMM's motion quality was generally found to be inferior to \textit{AAMDM}, as reflected in the FID and DIV metrics. This discrepancy is likely due to LMM's training with MSE loss, presupposing a one-to-one mapping. However, this assumption may not be valid in datasets with multiple possible transitions from a single pose. Figure~\ref{fig:diverse_example} provides an example. A more detailed discussion on this aspect will be presented in a subsequent section. However, LMM showed a higher FPS score, attributed to its single feed-forward operation, compared to \textit{AAMDM}'s five feed-forward operations.

In motion quality evaluation, MVAE slightly outperformed LMM with scores of 7.223 and 22.981 in DIV and FID respectively. MVAE's better quality can be linked to its use of VAE for handling multiple mappings in pose transitions. Although MVAE offered improved training stability and performance, \textit{AAMDM} still outperformed in these metrics. MVAE also exhibited faster performance than \textit{AAMDM} due to its single-step feedforward process.

In comparison between AMDM5 and \textit{AAMDM}, both methods used 5 diffusion steps which led to similar FPS scores (173 vs 192). However, the diffusion steps in AMDM5 were modeled using a Gaussian distribution, which is typically effective when the total number of denoising steps is in the order of hundreds. As AMDM5 utilized only five steps, this assumption did not hold and it led to compromised motion quality. On the other hand, \textit{AAMDM} leveraged DD-GANs to model multimodal transitions, which reduced the number of steps required for generating a new frame without sacrificing motion quality.

\begin{figure*}[t]
    \centering
    \includegraphics[width=0.85\textwidth]{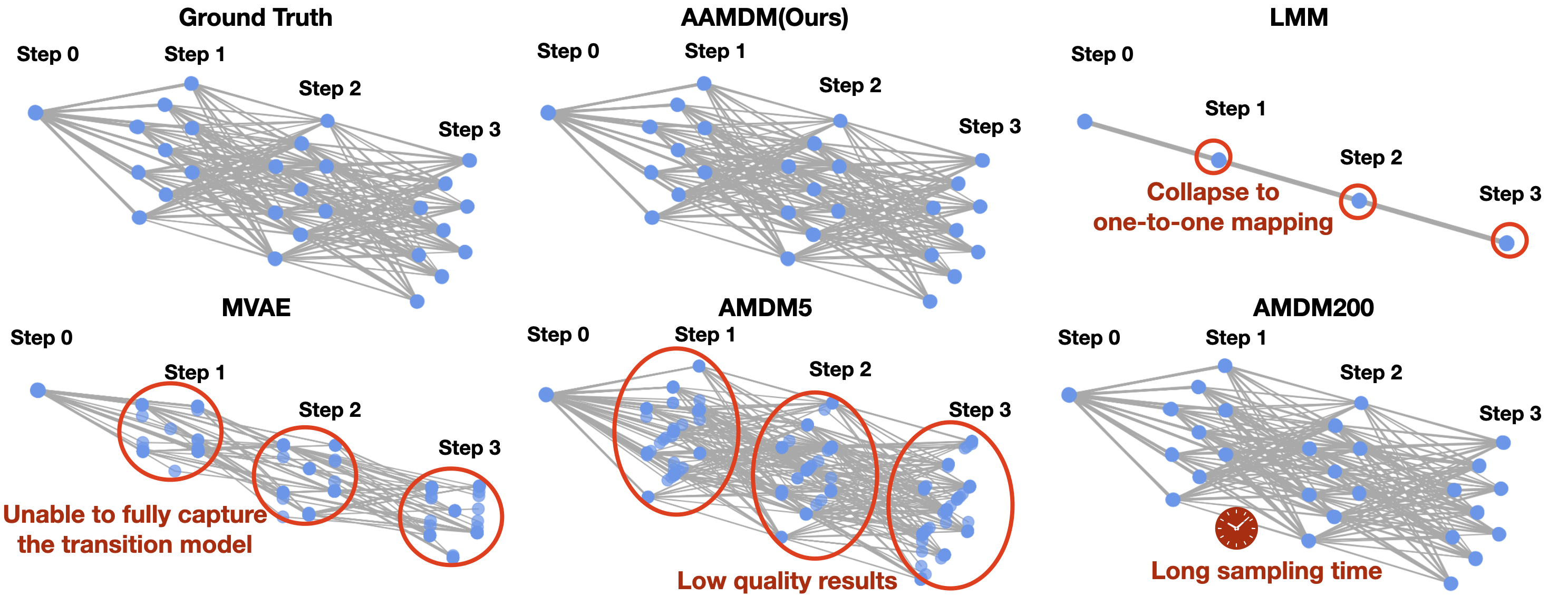}
    \vspace{-0.2cm}
    \caption{Visualization of the learned transition results of an artificial Squ-9-Gaussian experiment in 2D. We show that \textit{AAMDM} outperforms baseline methods in learning the many-to-many distribution mapping in sequential scenarios.}
    \label{fig: simple_exp_result}
    \vspace{-1em}
\end{figure*}

AMDM200 with more diffusion steps is better aligned with the Gaussian distribution assumption, which is connected to highly improved motion quality metrics. However, this increase in diffusion steps comes at the cost of efficiency. As the number of steps rises, the generation speed decreases. This trade-off highlights the balance between motion quality and generative efficiency, with AMDM200 favoring the former at the expense of the latter.

\subsubsection{Evaluation of Interactive Synthesis}

We evaluated the performance of these methods in an interactive motion synthesis scenario. The experiment involved interactively controlling the character's moving direction while allowing the arms to move freely. Our evaluation employed the following metrics, with '-UC' denoting 'Under Control':

\begin{itemize}
\item \textbf{Tracking Error (TE-UC)}: The TE-UC metric assesses the method's ability to follow user commands $\bar{\mathbf{x}}$. It is defined as the discrepancy between the user's command and the generated motion $|\bar{\mathbf{x}} - \hat{\mathbf{x}}|$.  A lower TE-UC value signifies better alignment with user input, reflecting superior performance.

\item \textbf{Frechet Inception Distance (FID-UC)}: The FID-UC is used to measure the similarity between the motion dataset and the generated trajectories. A lower FID-UC indicates a higher quality of the generated motion.

\item \textbf{Footskating Frame Ratio (FFR-UC)}: This metric evaluates the realism of the motion when the character is under user control. It assesses aspects such as naturalness and adherence to physical constraints. Lower FFR-UC scores suggest more physically plausible and realistic motion generation.
\end{itemize}

In user control scenarios, our results demonstrate that the \textit{AAMDM} framework consistently outperformed baseline methods across nearly all metrics evaluated. Compared with Learned Motion Matching (LMM), \textit{AAMDM} addresses several key issues inherent in the LMM's approach. LMM employs a projector network trained with an MSE loss to interpret user commands, which leads to two primary issues. Firstly, multiple candidate poses could potentially match the user command, but LMM's projector network struggles to handle multi-modal transitions. Secondly, the projector network often ignores the character's current pose, necessitating blending techniques to ensure smooth transitions. MVAE faces challenges in training to capture all the possible transitions, resulting in a quality of motion that does not match that of \textit{AAMDM}. Similarly, AMDM5's reduces the number of diffusion steps, which breaks the Gaussian distribution assumption and consequently downgrades the motion quality. Although AMDM200 provides higher-quality generation due to more diffusion steps, its low speed (4.72 FPS) is not suitable for any interactive applications.

\subsection{Additional Studies on Artificial Dataset}
In addition to the previous experiment, we conducted an additional study to analyze the effectiveness of various methods on a many-to-many transition dataset. 
For this purpose, we created a $2$D ``Squ-9'' dataset characterized by its multi-modal dynamics where any given point in three by three Gaussian distributions can transit to any other Gaussian distributions in the next time step. By learning this dataset, we evaluated the effectiveness of each method to capture this many-to-many dynamics. The comparative results are visually depicted in Figure~\ref{fig: simple_exp_result}.

In our results, our \textit{AAMDM} captured all the possible modes while preserving sample quality. In contrast, LMM struggled to represent the dataset's many-to-many vector transitions, resulting in a singular vector cluster at each step. MVAE showed an improvement in mode coverage, yet it cannot illustrate all possible modes. 
Among other diffusion model-based approaches, AMDM5 exhibited better transitions but their qualities are still worse than \textit{AAMDM}. Although AMDM200 produced results of comparable quality to \textit{AAMDM}, it required 40 times more inference time.

\begin{table}[h]
\centering
 \vspace{-0.4cm}

\begin{tabularx}{0.45\textwidth}{ c c | m{1.3cm} m{1.3cm} m{1.5cm} } 
\\ [0.5ex]
 \hline
 \textbf{$T^{ADM}$} & \textbf{$T^{GAN}$} & \textbf{DIV $\rightarrow$} & \textbf{FID $\downarrow$} & \textbf{FPS $\uparrow$}  \\ [0.5ex] 
 \hline 
\multicolumn{2}{c|}{Dataset} & 14.533   & 0.000  & N/A  \\[0.5ex] 
  \hline 
 0 & 3 & N/A  & N/A  & 311  \\ [0.5ex] 
 1 & 3 & 10.612  & 16.332 & 215\\ [0.5ex] 
 \;2$^*$ & \;3$^*$ &11.574  &14.051  & 173\\ [0.5ex] 
 10 & 3 &12.041  & 11.779 & 47\\ [0.5ex] 
 \hline
 2 &2 & 12.415  & 28.476  & 211\\ [0.5ex] 
 \;2$^*$ & \;3$^*$ &11.574  &14.051  & 173\\ [0.5ex] 
 2 & 4 & 11.313 & 14.534 &135 \\ [0.5ex] 
 2 & 10 & 9.775 & 16.312 &58 \\ [0.5ex] 
 \hline
\multicolumn{2}{c|}{wo/ Emb}  &56.341  &128.412  &146 \\ [0.5ex] 
 \hline
\end{tabularx}
\caption{Ablation study results. $^*$The default parameters.
}
\label{tab: ablation_studies}
\vspace{-0.4cm}
\end{table}

\subsection{Ablation Studies}
We provided additional insights of \textit{AAMDM} by conducting three ablation studies summarized in Table~\ref{tab: ablation_studies}.\\
\\
\textbf{Polishing Steps} In our study, we investigated the impact of the number of polishing steps ($T^{ADM}$) on the generation process. Specifically, we denote $T^{ADM}=0$ as the scenario where no polishing module is used, meaning the output from the generation module is directly utilized for future frame generation. In our experiments, settings with $T^{ADM}>0$ exhibited significant performance enhancements compared to the $T^{ADM}=0$ scenario as when $T^{ADM}=0$, the framework was unable to generate reliable long-horizon trajectory due to the diverges of the character's pose. This suggests that relying solely on denoising diffusion GANs may not yield high-quality outputs for long-horizon generation. In contrast, additional polishing steps markedly improved the output quality, making it more suitable for long-horizon predictions. Furthermore, results indicate a positive correlation between the number of polishing steps $T^{ADM}$ and the output quality. However, it is important to note that increasing $T^{ADM}$ also leads to longer sampling times.\\
\\
\textbf{Generation Steps} In our second study, we examined the effects of the number of generation steps. Theoretically, increasing $T^{GAN}$ should reduce the amount of noise that needs to be removed at each denoising step, potentially simplifying the training process. However, our results show that a specific value of $T^{GAN}$, $T^{GAN}=3$ and $T^{GAN}=4$, yielded the highest overall motion quality,  yet  $T^{GAN}=3$ was more efficient. Although $T^{GAN}=2$ achieved the best performance in the DIV metric, we observed a few cases of divergence in the motion, which resulted in worse performance in FID compared to $T^{GAN}=3$. When $T^{GAN}=10$, the learning task should be easier since the distance between each diffusion step is smaller. However, our results show that the performance was the worst. We hypothesize that this is because we utilized a simple MLP network; thus, it may not be adequately equipped to handle larger values of $T^{GAN}$ for effectively training the denoising diffusion GANs.\\
\\
\textbf{Importance of Embedded Transition Space} In this analysis, we explored the advantages of learning transitions in an embedded space $\mathbf{XZ}$ as opposed to the full pose space $\mathbf{Y}$. Our results suggest that utilizing the full pose space yields inferior outcomes compared to an embedded space. We attribute this finding to two primary factors. Firstly, learning in a higher-dimensional space, like the full pose space, is inherently more challenging than in a lower-dimensional space, particularly under multimodal distribution conditions. Secondly, as discussed in the previous section, \textit{AAMDM} does not employ a complex neural network architecture or specialized techniques for constructing the latent space in Denoising Diffusion GANs. MLP networks used in our framework may not be sufficiently robust to capture transitions in larger spaces effectively. This limitation further supports the advantage of using an embedded space for learning transitions.

%% file: paper_text/disscusion_conclusion.tex
\section{Discussion and future work}
We have introduced a novel framework for motion synthesis: Accelerated Auto-regressive Motion Diffusion Model (\textit{AAMDM}). \textit{AAMDM} is designed to efficiently generate high-quality animation frames for interactive user engagement. This is achieved by several technical components: the use of a low-dimensional embedded space for compact representation, Denoising Diffusion GANs for fast approximations, and the Diffusion Model for robust and accurate long-horizon synthesis. Our benchmarking of \textit{AAMDM} against various baseline methods has demonstrated its superior capabilities in motion synthesis. We have also investigated the nuances of different autoregressive motion synthesis methods, providing valuable insights into this domain. Additionally, our ablation studies have validated the design choices made for \textit{AAMDM} and identified the influence of various hyperparameters on the overall system performance.

In the future, we plan to explore several research directions. 
One notable challenge is the trade-off between the motion quality and the computational cost.
Future work could explore advanced techniques such as parallel computing and the use of temporal information to accelerate the generation process.
In addition, the model performance can be further improved by introducing more sophisticated methods to structure the latent space of the denoising diffusion GANs, such as a structured matrix-Fisher distribution~\citep{sengupta2021hierarchical}. 
Finally, it will be interesting to improve the controllability of the framework by introducing a learning-based control mechanism rather than relying on gradient-based sampling guidance.

%% file: paper_text/appendix.tex
\clearpage
\setcounter{page}{1}
\maketitlesupplementary

\section{Training Detail}
In this section, we provide extra information for training \textit{AAMDM}.
\subsection{Training Procedure}
\input{algos/DM3_training}

\subsection{Training Hyperparameter}
Here we list the training hyperparameter we use:
\begin{itemize}
    \item Learning rate for $G^{ADM}$: 3e-4.
    \item Learning rate for $G^{GAN}$: 1e-4.
    \item Learning rate for $D^{GAN}$: 1e-4.
    \item Training windows size h: 10.
    \item Batch Size: 64.
    \item Noise scheduling: [3.764e-4, 1.452e-3, 0.257, 0.668, 0.999] 
\end{itemize}

\begin{figure}
    \centering
    \includegraphics[width=0.495\textwidth]{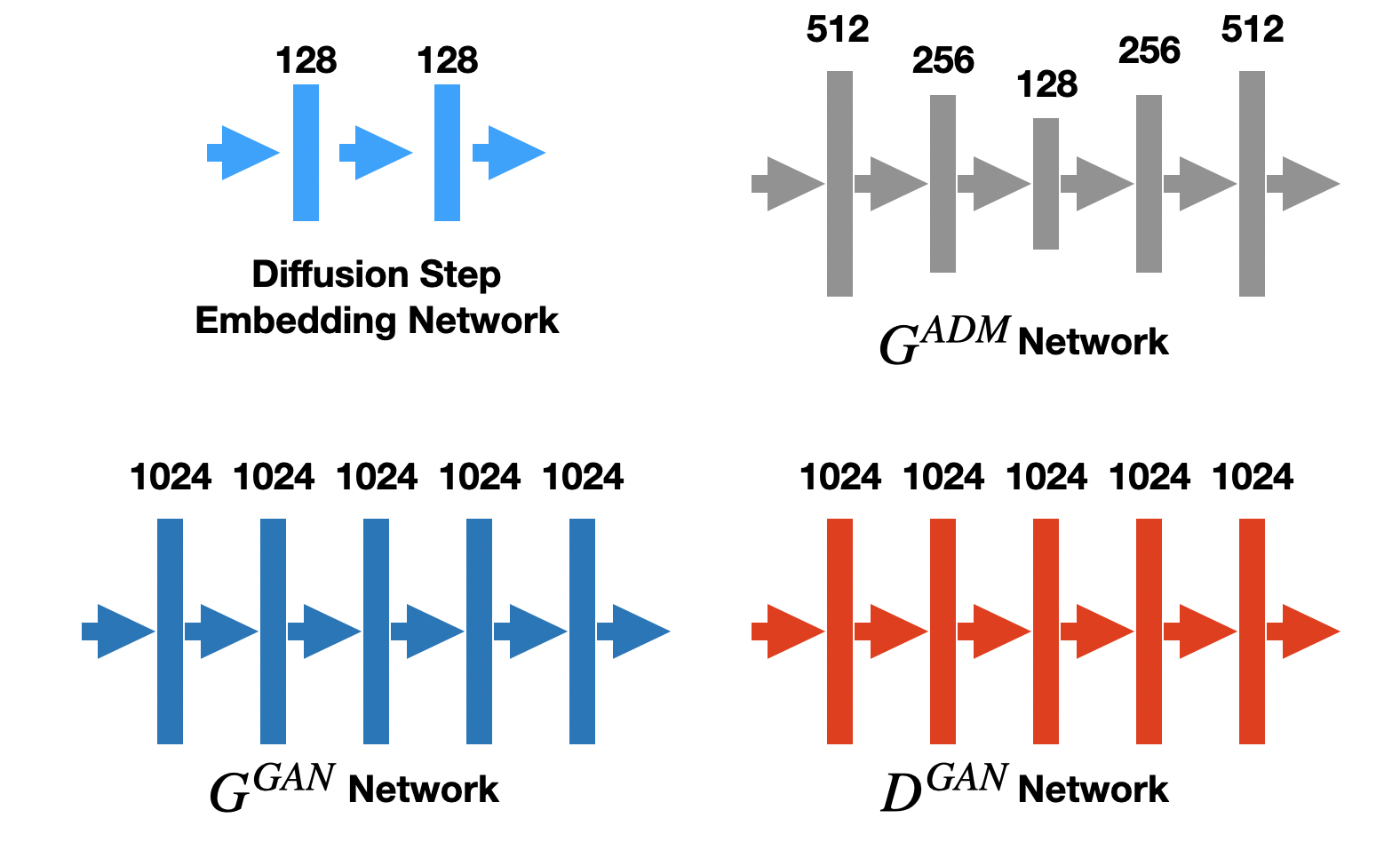}
    \caption{\small{The network structure used in AAMDM. We use Mish as activation function for all networks.}}
    \label{fig:networks}
\end{figure}

The neural network structure of each module are shown in Figure:~\ref{fig:networks}.

%% file: algos/DM3_training.tex
\begin{algorithm}[H]
  	\caption{AAMDM Learning pseudo-code}
  	\label{alg:AAMDM_Learning}
  	\begin{algorithmic}[1]{\small

          \Require  Embedded Vector Dataset $\mathbf{XZ}$, Forward Diffusion $FD$, Noise Factor $\alpha$, Total Diffusion Steps $T^{AA}$, Diffusion Steps in Polishing Module $T^{ADM}$
          \item[]

        \State Intialize: Generator Module $G^{GAN}$, Polishing Module $G^{ADM}$, Discriminator Network $D^{GAN}$
        \Repeat
        \State Sample $\bfx\bfz$ trajectory from $\mathbf{XZ}$:  $\bfx\bfz_{0:h}$
        
        \item[]
        \State {{\textsc{// Roll-out Polishing Module}}}
        \State Initialize n $\leftarrow$ 1
        \State Initialize $\hat{\bfx\bfz}^0_{n-1} \leftarrow \bfx\bfz_{k-1}$ 
        \Repeat
        \State Sample Polishing Module step
        $t \sim [1, T^{ADM}]$
        \State Forward diffusion $\bfx\bfz^t_{n} \leftarrow FD(\bfx\bfz_{n})$
        \State Reverse diffusion $\hat{\bfx\bfz}^0_{n} \leftarrow G^{ADM}(\bfx\bfz^t_{n},\hat{\bfx\bfz}^0_{n-1}, t )$
        
        \State $n \leftarrow n+1$
        \Until{n==h}
        \item[]
         \State {{\textsc{// Update Models}}}
        \State Update $G^{ADM}$ using  $L^{ADM}(\bfx\bfz_{1:h}, \hat{\bfx\bfz}^0_{:h})$
         
        \State Sample Generation Module step
        $t \sim [T^{ADM}, T^{AA}-1]$
        \State Forward diffusion $\bfx\bfz^{t,real}_{1:h} \leftarrow FD(\hat{\bfx\bfz}^0_{1:h})$
        \State Forward diffusion $\bfx\bfz^{t+1}_{1:h} \leftarrow FD(\bfx\bfz^t_{1:h})$
        \State Sample $r^{t+1} \sim N(0, I)$
        \State  \small{$\hat{\bfx\bfz}^0_{k} \leftarrow G^{GAN}(\bfx\bfz^{t+1}_{k},\hat{\bfx\bfz}^0_{k-1}, r^{t+1}, t+1)$} for k in [1,h]
        \State Forward diffusion $\bfx\bfz^{t,fake}_{1:h} \leftarrow FD(\hat{\bfx\bfz}^0_{1:h})$
        \State Update $G^{GAN}$, $D^{GAN} $using Equation \ref{equ:diffgan_gen} and  \ref{equ:dis_loss}.
        
        \Until{Converge}

        }
  	\end{algorithmic}
\end{algorithm}